\documentclass[letterpaper,twocolumn,10pt]{article}
\usepackage{usenix}

\usepackage{tikz}
\usepackage{amsmath}

\usepackage{filecontents}
\usepackage{graphicx}
\usepackage{subcaption}

\DeclareRobustCommand{\whitecircled}[1]{%
  \tikz[baseline=(char.base)]{%
    \node[shape=circle,draw,fill=white,inner sep=1pt]
         (char){\textcolor{black}{\footnotesize #1}};}}

\begin{document}

\date{}

\title{\Large \bf xGR: Efficient Generative Recommendation Serving at Scale}


\author{
{\rm Qingxiao Sun}$^{1}$,
{\rm Tongxuan Liu}$^{2}$,
{\rm Shen Zhang}$^{1}$,
{\rm Siyu Wu}$^{1}$,
{\rm Peijun Yang}$^{2}$,
{\rm Haotian Liang}$^{3}$, 
{\rm Menxin Li}$^{2}$, \\
{\rm Xiaolong Ma}$^{2}$, 
{\rm Zhiwei Liang}$^{2}$, 
{\rm Ziyi Ren}$^{2}$,
{\rm Minchao Zhang}$^{2}$,
{\rm Yifan Wang}$^{2}$,
{\rm Xinyu Liu}$^{4}$, 
{\rm Ke Zhang}$^{2}$, \\
{\rm Hailong Yang}$^{1}$, 
{\rm Depei Qian}$^{1}$
\\[1ex]
$^1$ Beihang University \quad
$^2$ JD Company \quad
$^3$ University of Science and Technology Beijing \quad
$^4$ Huawei
}

\maketitle

\begin{abstract}
Recommendation system delivers substantial economic benefits by providing personalized predictions. Generative recommendation (GR) integrates LLMs to enhance the understanding of long user-item sequences. Despite employing attention-based architectures, GR's workload differs markedly from that of LLM serving. GR typically processes long prompt while producing short, fixed-length outputs, yet the computational cost of each decode phase is especially high due to the large beam width. Furthermore, since the beam search involves a vast item space, the sorting overhead becomes particularly time-consuming. We propose xGR, a GR-oriented serving system that meets strict low-latency requirements under high-concurrency scenarios. First, xGR unifies the processing of prefill and decode phases through staged computation and separated KV cache. Second, xGR enables early sorting termination and mask-based item filtering with data structure reuse. Third, xGR reconstructs the overall pipeline to exploit multi-level overlap and multi-stream parallelism. The experiments on real-world datasets demonstrate that xGR achieves at least 2.89$\times$ throughput compared to the state-of-the-art baseline under strict latency constraints.


\end{abstract}

\section{Introduction}
\label{sec:intro}

Recommendation systems play a central role in modern digital platforms, such as social media~\cite{huang2017picture}, video~\cite{covington2016deep}, music~\cite{fayyaz2020recommendation}, and e-commerce~\cite{gu2021self} applications. By providing personalized and accurate predictions, recommendation systems can enhance user engagement and substantially improve economic benefits. For example, YouTube’s recommendation system has helped over 1 billion
users discover personalized videos, resulting in a 60\% increase in clicks~\cite{covington2016deep}. As users generate increasingly rich behavioral data across various application scenarios, the demand for intelligent recommendation systems among enterprises continues to rise~\cite{ko2022survey}.


The success of industrial applications has promoted the development of recommendation systems~\cite{zangerle2022evaluating}. During the past decade, deep learning-based recommendation (DLR)~\cite{guo2017deepfm,cheng2016wide} has gained widespread attention. DLR typically employs a cascade architecture~\cite{ibrahim2025revisiting}, which includes sequential stages such as recall, pre-ranking, and fine-ranking. Among these, fine-ranking is the most computation-intensive stage, which utilizes \textit{deep learning recommendation models (DLRMs)}~\cite{liu2022monolith,naumov2019deep} to predict the click-through rate (CTR) and conversion rate (CVR) scores. However, such a cascade architecture leads to system fragmentation, with a significant amount of time consumed in inter-stage communication, making it difficult to meet the ever-increasing service level objective (SLO) requirements~\cite{ye2023grace,wang2024atrec}. Furthermore, DLRM generally adopts simple MLP networks, heavily relying on feature engineering and offering limited understanding of user behavior~\cite{liu2022monolith}. 

In recent years, \textit{generative recommendation (GR)} systems have experienced explosive growth~\cite{lin2025can}. Unlike traditional DLR, GR incorporates a large language model (LLM), leveraging its rich pre-trained corpus to mitigate the cold-start problem for new items~\cite{deldjoo2024review}. Furthermore, the inherent autoregressive nature of LLMs enables GR to follow a robust scaling law~\cite{han2025mtgr,wang2025scaling}, and the strong long-sequence embedding capability makes it particularly well-suited for extended user context in recommendation scenarios. In early development, LLMs are exploited to collaborate or replace certain stages of DLR cascade architecture, such as recall strategy enhancement~\cite{rajput2023recommender} and fine-ranking model upgrades~\cite{zhai2024actions}. Subsequently, LLMs serve as the core of recommendation systems~\cite{deng2025onerec}, which enable end-to-end recommendation through generative paradigms. This not only avoids error propagation among stages but also creates opportunities for deploying recommendation functionality directly via LLM inference.


Although GR employs an attention-based architecture similar to general LLMs~\cite{vaswani2017attention,liu2024deepseek,achiam2023gpt}, its inference workload characteristics differ significantly from LLM serving due to the unique nature of recommendation scenarios. 
At first, GR involves a fixed number of decode phases~\cite{du2025prefillonly}, but each decode incurs substantial computation, requiring the avoidance of redundant KV cache loading and additional block copying.
In addition, GR needs to perform efficient beam search~\cite{xie2023self} within a vast item space and coordinate resources for item filtering throughout the overall generation process.
However, the core bottleneck lies in the extremely low latency requirement of the entire online inference pipeline (within tens to 200 milliseconds), while peak user traffic reaches thousands of queries per second (QPS)~\cite{yang2025gpu}. 
These characteristics suggest that naively adopting an LLM inference system is inadequate for GR, and a comprehensive reconstruction of the underlying pipeline is required. Specifically, the GR system exhibits three unique challenges that motivate this paper.

\textit{Challenge 1: how to minimize the computation and memory cost induced by GR's long prompt, short fixed-length output nature?} Unlike conversational LLM applications that generate long responses to short prompts~\cite{zhong2024distserve}, the GR task is characterized by long prompt inputs and fixed-length, short outputs. For example, a user's lifetime behavior sequence may contain hundreds or thousands of tokens, but the output requires only a few tokens. 
This offers the potential for aggressive prefill-only optimizations~\cite{du2025prefillonly}. But in GR, the computational volume of each decode phase becomes especially large due to the incorporation of beam search. Also, existing methods generally treat each beam sequence as independent~\cite{kwon2023efficient,liu2025xllm}, ignoring the fact that they actually share an identical prompt context. The redundant loading of the shared KV cache exacerbates the memory bandwidth bottleneck. On the other hand, when the input length does not align with the KV block size, the last block has to be explicitly copied to ensure the continuity of beam sequences. The costly massive block copies introduce severe latency and fragmented memory allocations.



\textit{Challenge 2: how to maintain beam search efficiency while minimizing both sorting and filtering overhead in high-concurrency recommendation scenarios?} The core operation of beam search is to select beam width ($BW$) sequences from a large pool of candidate sequences~\cite{meister2020if}. 
For the identified tokens, only the Top-$K$ likely candidate tokens for the next step are retained~\cite{leblond2021machine}. Accurate recommendations require larger $BW$ and $K$ values, but the sorting and filtering operations become particularly time-consuming. As the decode phase progresses, beam search continuously generates new candidate sequences and discards old sequences. Frequent creation and destruction of related data structures incur significant overhead. 
Moreover, due to the numerous possibilities of token ID combinations, not all of them correspond to actual products, leading to candidate sequence invalidation~\cite{kong2025minionerec}. 

\textit{Challenge 3: how to schedule multi-level serving pipeline to maximize parallelism while reducing accelerator idle time?} The scheduler first prepares the information needed on the host side. After preparation, the relevant tasks are handed over to the engine. The engine executes one prefill as well as several beam search and decode combinations~\cite{deng2025onerec}. The workers are responsible for executing specific phases. In such a multi-level pipeline, the host and device handle distinct tasks, and the unnecessary global synchronization leads to serial execution. In addition, request sizes typically follow a power-law distribution that potentially contains tens to thousands of tokens. Moreover, current GR models~\cite{yang2025qwen3} contain far fewer parameters than that of LLMs, causing host-side scheduling to account for a substantial portion of the latency. 




To address the above challenges, we propose xGR, a serving system that meets the low-latency constraints under high-concurrency scenarios. \textit{For Challenge 1,} xGR achieves efficient memory management and load balancing through KV cache separation and staged attention computation. \textit{For Challenge 2,} xGR minimizes search overhead with early sorting termination and mask-based item filtering.
\textit{For Challenge 3}, xGR leverages host-device overlap and multi-stream parallelism in the system pipeline.
xGR has been deployed in production scenarios for over 6 months, serving over hundreds of millions users. Even under peak RPS of tens of thousands, it maintains a P99 latency within 200 milliseconds.
To the best of our knowledge, this is the first work to comprehensively reconstruct the GR serving paradigm, encompassing optimizations from operator and algorithm to system.  Specifically, this paper makes the following contributions:



\begin{itemize}

\item We comprehensively analyze the workload characteristics of the GR paradigm and expose potential performance optimization opportunities to facilitate deployment.

\item We optimize the overall process spanning prefill, decode, and beam search phases. At the operator level, xAttention accelerates inference phases through KV cache management and staged attention computation. At the algorithm level, xBeam reduces search overhead with valid path constraint and early search termination.


\item We deconstruct the serving pipeline to facilitate system-level optimizations. xSchedule employs a three-tier hierarchy to enable pipeline overlap and task parallelism.


\item We develop a GR serving system xGR that enables fast recommendations under high-concurrency request scenarios. The experimental results show that xGR achieves at least 2.89$\times$ throughput compared to the state-of-the-art baseline under strict P99 latency contraints.

\end{itemize}









\section{Background}
\label{sec:background}


\subsection{Recommendation Paradigms} 

%




\subsubsection{Discriminative Recommendation}

To handle massive item pools, traditional industrial recommendation systems universally adopt a discriminative cascade architecture \cite{ibrahim2025revisiting}. This pipeline typically consists of three sequential filtering stages: \textit{Recall}, which selects thousands of candidates from hundreds of millions of items using two-tower models or inverted indices, focusing on coverage; \textit{Pre-ranking}, which uses lightweight models to narrow down the scope; and \textit{Fine-ranking}, the most computationally intensive stage, which typically employs DLRM\cite{liu2022monolith,naumov2019deep} to predict precise CTR/CVR scores for hundreds of candidates. However, this architecture faces two fundamental limitations. First, the I/O overhead and network transmission accumulated across multiple model instances introduce significant system latency. Second, sparse embedding lookups (\textit{gather} operations) dominate execution time, resulting in low arithmetic intensity. 
\subsubsection{Generative Recommendation}

\begin{figure}[htbp]
    \centering
    \includegraphics[width=1.0\linewidth]{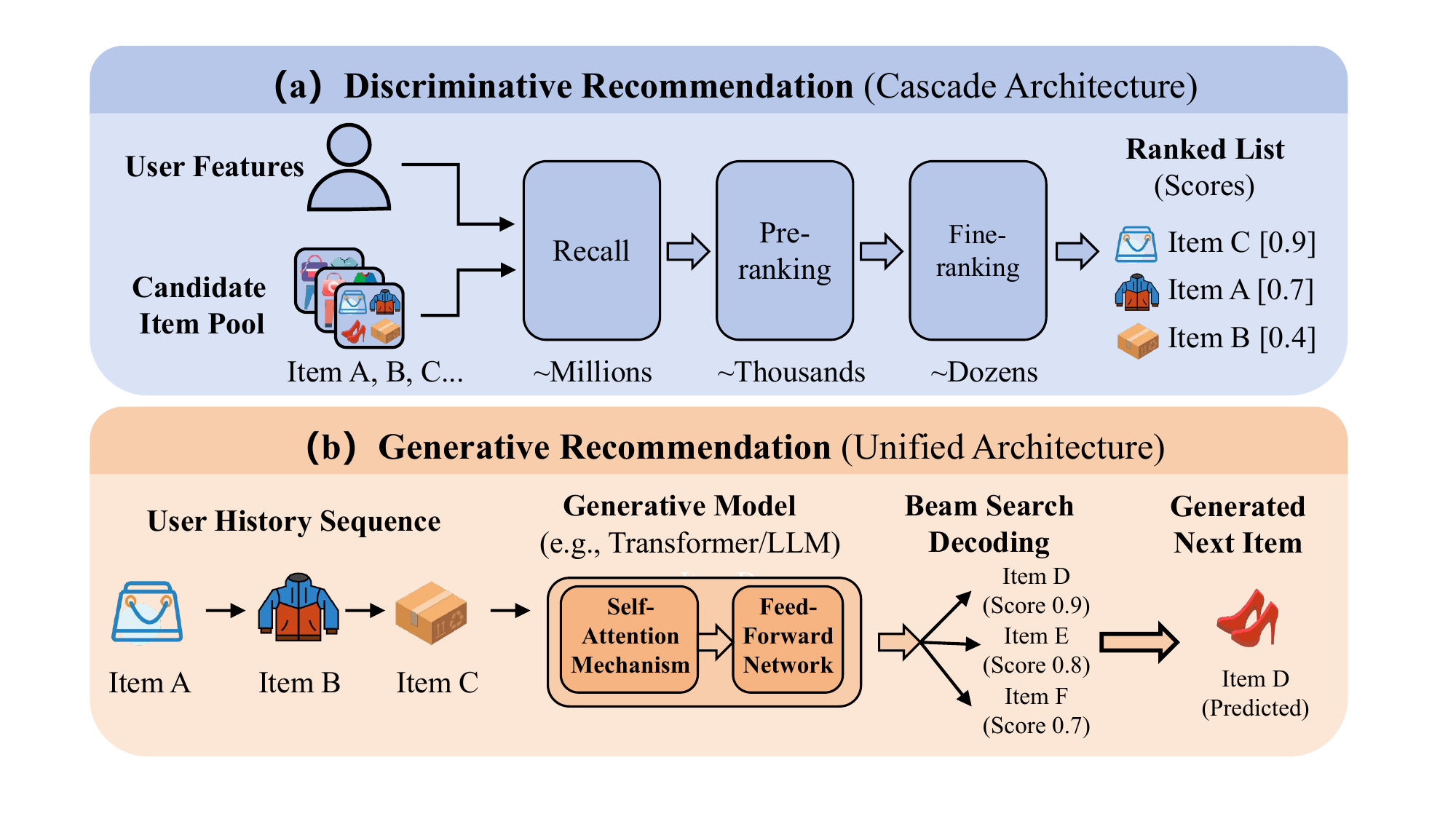}
    \caption{Architecture comparison between discriminative recommendation and generative recommendation.}
    \label{fig:gr}
\end{figure}

To address these limitations, the industry pivots towards GR. As shown in Figure \ref{fig:gr}, departing from the multi-stage discriminative architecture, GR adopts a Transformer-based unified architecture (e.g., TIGER \cite{rajput2023recommender}, OneRec \cite{deng2025onerec}). GR redefines the recommendation task as a \textit{Sequence-to-Item} generation problem: given a user's historical behavior sequence $H = \{x_1, x_2, ..., x_t\}$, the model directly predicts the next most probable item $x_{t+1}$.

The core driver behind this shift is the scaling law \cite{kaplan2020scaling, han2025mtgr,wang2025scaling}. Recent studies indicate that the accuracy of traditional DLRM tends to saturate at a certain scale, whereas Transformer-based GR exhibits strong scalability—recommendation quality improves predictably as model parameters, data volume, and FLOPs increase. 



\subsection{System Analysis of GR}

\subsubsection{GR Inference Workflow}

GR follows the standard ``Prefill-Decode'' paradigm but exhibits unique characteristics.


\textbf{Prefill.} In contrast to LLMs that typically process concise prompts, GR encodes a significantly longer sequence of user interaction history (e.g., extensive lists of clicked item IDs), resulting in a long prompt input workload.

\textbf{Beam Search \& Decode.} While LLMs typically employ stochastic sampling to foster diversity and creativity~\cite{li2019robust}, GR prioritizes retrieval accuracy to recommend the most relevant items. 
Unlike open-ended text generation, stochastic sampling is not suitable for recommendation scenarios that return a small set of top-ranked items. Also, random decoding can easily miss high-probability candidates and even produce semantic-ID combinations that do not correspond to real items~\cite{deng2025onerec,kong2025minionerec,lin2024efficient}.
To this end, GR adopts beam search algorithm to balance the trade-off between retrieval quality and computational cost. It maintains a set of $BW$ most promising partial sequences, where $BW$ is the beam width. 

\begin{figure}[htbp]
    \centering
    \includegraphics[width=1.0\linewidth]{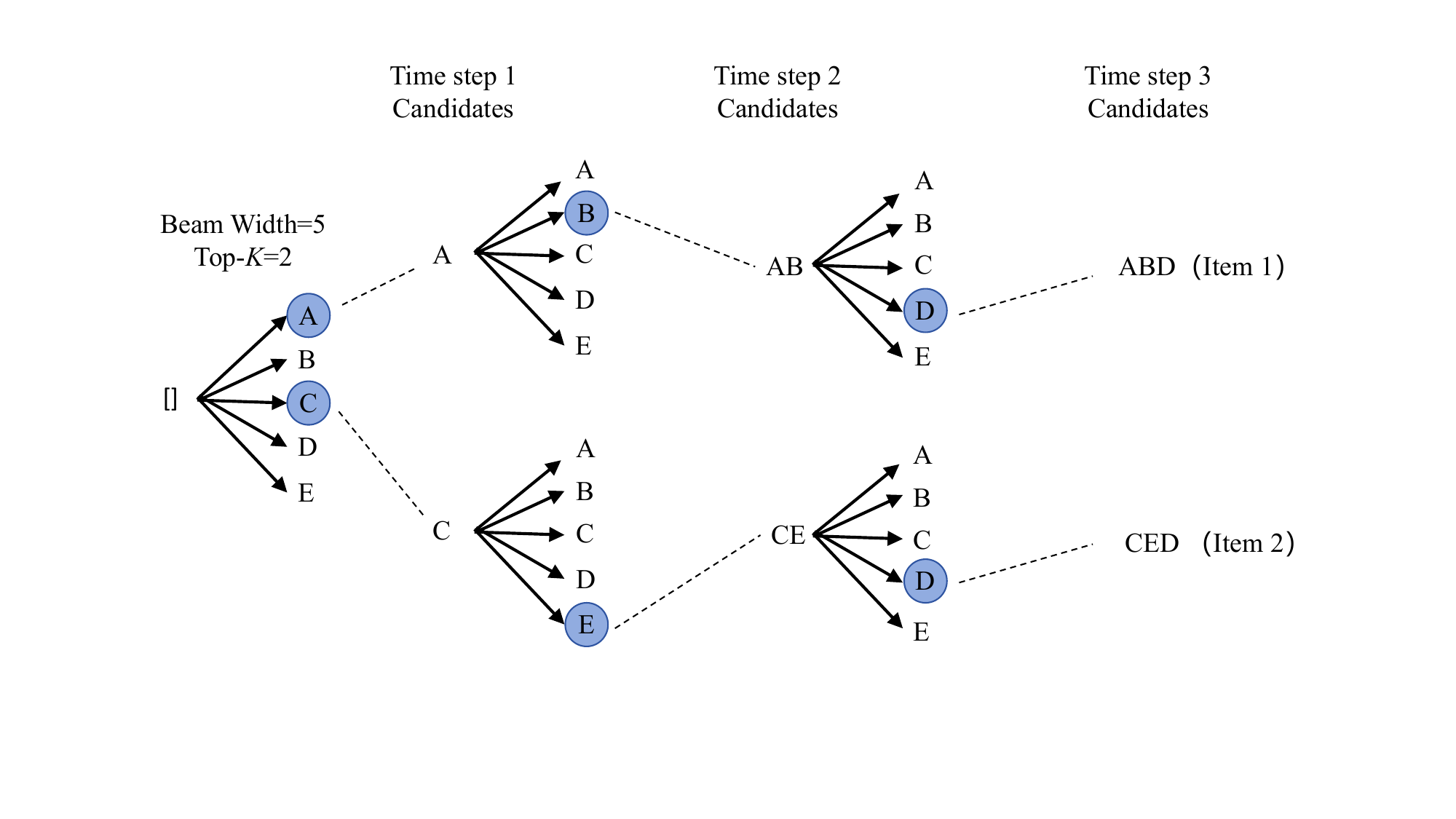}
    \caption{An example of beam search process. Each capital letter denotes one semantic ID token, and three tokens together represent one item.}
    \label{fig:beam}
\end{figure}

As shown in Figure \ref{fig:beam}, at each time step $t$, the algorithm expands all active beams by computing the probability distribution over the item vocabulary. To manage the search space, the system first selects the $BW$ most likely next-token candidates for each beam. From this aggregated pool of $BW \times K$ candidates, it then identifies the global Top-$K$ sequences with the highest cumulative log-probabilities to form the new set of active beams for step $t+1$. Since each item is represented by a semantic ID composed of three token IDs, so the process terminates after these time steps. Since recommendation scenarios demand high diversity, large $BW$ and $K$ values (e.g., $128 \times 128$ or even $512 \times 512$) are typically required. This results in a massive candidate pool (up to $2.6 \times 10^5$ candidates per step), transforming the lightweight sorting operation into a computationally heavy bottleneck. Furthermore, it forces the system to maintain and update large amount of active beams simultaneously at each step, rendering the GR decode phase significantly different from the single-stream decoding commonly found in LLM applications.

\subsubsection{Key Workload Characteristics}
\label{sec:background:char}
Based on observations from production environments, GR inference workloads exhibit three unique characteristics in current LLM systems.


\textbf{Redundant memory access in long-context beam search.} In GR, a single long user history sequence (prompt) is shared across a wide beam width ($>=$128) to generate diverse item candidates. Existing mechanisms natively treat these beams as independent sequences during decode calculations, necessitating repetitively loading the massive prompt KV cache from global memory to compute units for each beam. 


\textbf{Memory inefficiency from beam forking.} The dynamic nature of beam search, where sequences continuously fork and retire, introduces severe inefficiencies in block-based memory management (e.g., PagedAttention~\cite{kwon2023efficient}). If the sequence length does not align with the KV block size, the system is forced to perform physical block copies to ensure context independence for the new beam branches. 

\textbf{Latency sensitivity due to strict SLO and small model scale.} GR services operate under extremely strict SLO, typically requiring P99 latency to be within 200 milliseconds to ensure user experience. Unlike general LLMs with hundreds of billions of parameters, GR models are often relatively smaller (e.g., 100M to 10B parameters). This shift in model scale makes the system highly sensitive to non-compute overheads such as host-side scheduling and synchronization. 



\subsection{Accelerator Abstraction}
\label{sec:accelerator}

Modern accelerators typically comprise dozens of core groups (CGs) that share an L2 cache through the interconnect network. The L2 cache is connected to high-bandwidth global memory. Each CG consists of different functional units, including matrix compute units (MCUs), vector compute units (VCUs), and scalar compute units (SCUs).
The register file serves as the storage closest to these compute units, followed by L1 cache and scratchpad memory. The scratchpad memory is explicitly managed, thus offering high flexibility.




\begin{table}[htbp]
\caption{Correspondence in Ascend NPU and NVIDIA GPU.}
\renewcommand{\arraystretch}{1.2}
\centering
\footnotesize
\begin{tabular}{c||c|c}
\hline
Abstraction       & Ascend GPU     & NVIDIA GPU    \\ \hline \hline
Core Group              & AI Core        & Streaming Multiprocessor            \\ \hline
Matrix Compute Unit              & Cube           & Tensor Core   \\ \hline
Vector Compute Unit              & Vector Unit    & CUDA Core     \\ \hline
Scalar  Compute Unit            & Scalar Unit    & —             \\ \hline
Scratchpad Memory & Unified Buffer & Shared Memory \\ \hline
\end{tabular}
\label{tab:hardware}
\end{table}

Table~\ref{tab:hardware} presents the correspondence between the unified abstraction and the architectures of Ascend NPU and NVIDIA GPU. For such a unified abstraction, performance optimizations must pay particular attention to the following principles: \whitecircled{1} the complex storage hierarchy necessitates careful data placement; \whitecircled{2} the diversity of compute units necessitates fine-grained task partitioning; \whitecircled{3} task dependencies necessitate efficient execution pipelines. These principles guide us in designing specialized optimizations for GR workloads.


\section{Motivation}
\label{sec:motivation}

\subsection{Attention Performance}
\label{sec:performance}



\begin{figure}[htbp]
    \centering
    \includegraphics[width=0.9\linewidth]{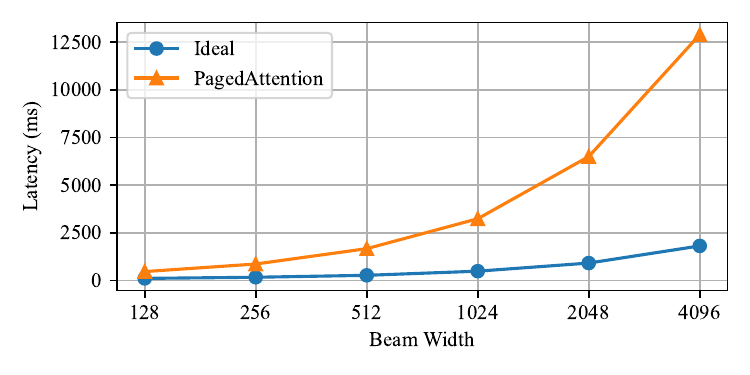}
    \caption{Latency comparison between PageAttention and Ideal implementation.}
    \label{fig:performance}
\end{figure}

Existing attention kernels lack awareness of the shared prefix inherent in beam search, resulting in significant memory access overhead. As shown in Figure \ref{fig:performance}, the latency of PagedAttention~\cite{kwon2023efficient} exhibits a significant upward trend as the beam width increases. In contrast, the Ideal that represents theoretical performance with perfect reuse of the shared prefix remains relatively flat. The significant performance gap between the actual curve of PagedAttention and Ideal intuitively reveals the substantial potential for optimization by eliminating redundant memory access to enhance operator efficiency. 


\subsection{Memory Consumption}
\label{sec:memory}

\begin{figure}[htbp]
    \centering
    \includegraphics[width=0.9\linewidth]{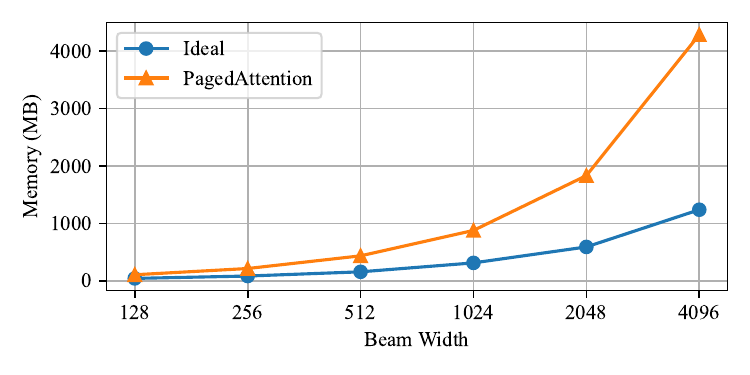}
    \caption{Memory consumption comparison between PageAttention and Ideal implementation.}
    \label{fig:memory}
\end{figure}

The paged memory management mechanism introduces significant copy overhead and memory redundancy when handling beam forking. As shown in Figure \ref{fig:memory}, with the increase in beam width, the memory consumption of PagedAttention increases sharply due to frequent block copies and resulting fragmentation. In contrast, the Ideal implementation stores only one copy of the shared prefix in the memory.
The large gap between PagedAttention and the Ideal confirms that its memory management requires urgent optimization. 




\subsection{Invalid Item Generation}
\label{sec:valid}

\begin{figure}[htbp]
    \centering
    \includegraphics[width=0.9\linewidth]{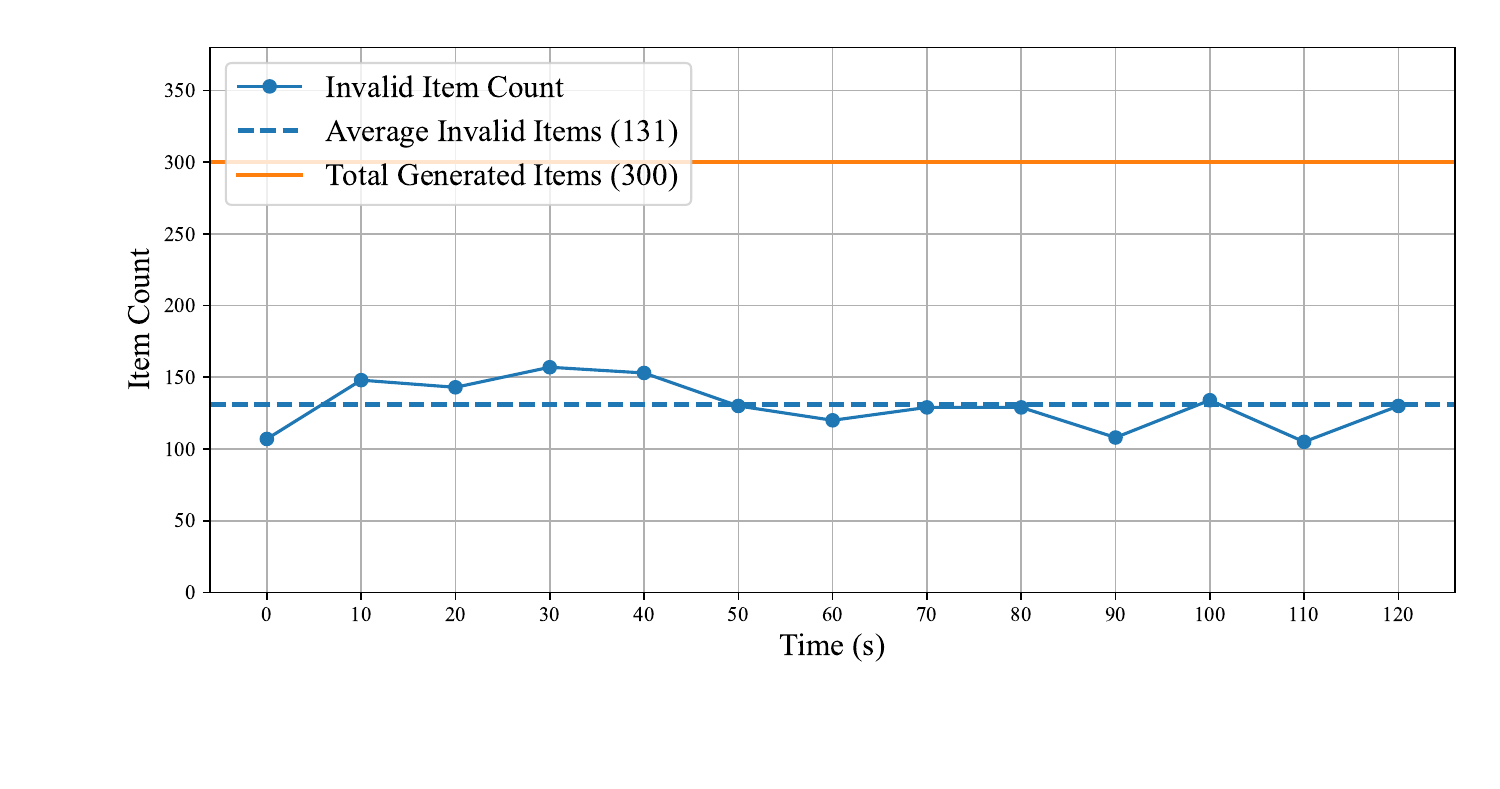}
    \caption{Proportion of invalid items under the total generation capacity of 300 items within a 2-minute interval.}
    \label{fig:item}
\end{figure}

The fundamental difference between GR and general text generation lies in the validity constraint of the output. Given the exponentially growing combination space of Token IDs, not all token sequences map to valid, real-world items. As shown in Figure \ref{fig:item}, the experiments conducted without filtering conditions reveal that the proportion of invalid items approaches 50\%. These invalid results not only waste compute resources but also significantly degrade the final recommendation quality
However, existing mechanisms for item filtering face a severe dilemma: calculating the masks dynamically introduces substantial latency, while pre-storing the masks leads to unmanageable memory overhead.



\section{Overview}
\label{sec:overview}

\begin{figure}[htbp]
    \centering
    \includegraphics[width=1.0\linewidth]{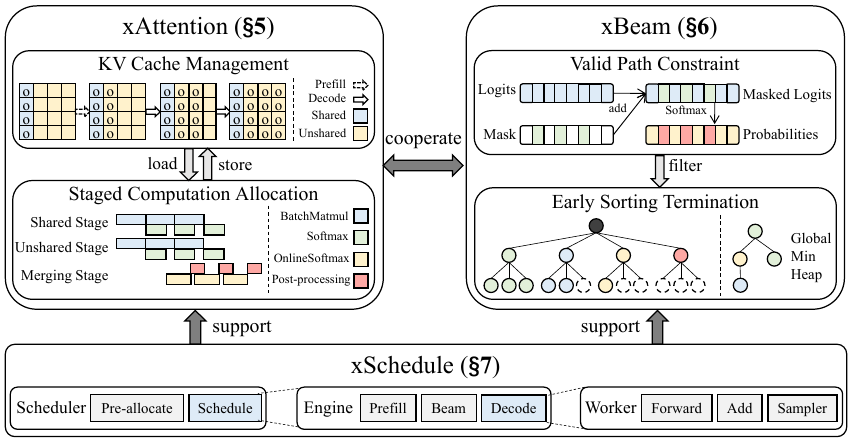}
    \caption{Design overview of xGR.}
    \label{fig:overview}
\end{figure}

In this section, we present xGR, a GR-oriented serving system designed to meet strict SLO requirements under high request concurrency. As shown in Figure~\ref{fig:overview}, xGR is composed of three tightly coupled components: xAttention for operator-level optimizations (\S5), xBeam for algorithm-level optimizations (\S6), and xSchedule for system-level optimizations (\S7).

Figure~\ref{fig:overview} illustrates the interaction of the three components. xAttention and xBeam span the prefill, decode, and beam phases, working cooperatively to achieve efficient token generation and sampling. Specifically, xAttention focuses on self-attention computation, integrating KV cache management and staged computation allocation to improve throughput. xBeam quickly determines the candidate tokens via valid path constraint, early sorting termination, and data structure reuse. xSchedule provides system support for xAttention and xBeam. It employs a three-tier management hierarchy that enables pipeline overlap and task parallelism across batches, across requests within a batch, and within each request.
A detailed description is provided in the following sections.

  
\section{Attention Computation}
\label{sec:attention}

Attention computation remains the primary performance bottleneck in GR scenarios. Specifically, the process is divided into one prefill phase and several decode phases, with each decode phase generating a token ID ($TID$), where the resulting $TID$ triplet represents an item ID. Unlike prior study that identifies the prefill phase as the dominant cost~\cite{du2025prefillonly}, large $BW$ and $K$ settings make the decode stage also particularly time-consuming. In the following, we present the KV cache management and task allocation mechanisms.

\subsection{KV Cache Management}
\label{subsec:kv}

Existing KV cache management strategies are highly inefficient in GR scenarios. For example, PagedAttention~\cite{kwon2023efficient} ignores the fact that sequences within a beam share the same prefix token IDs, and instead redundantly loads identical KV blocks for expansion. 
Although recent works~\cite{zhu2024relayattention,shyam2024tree,lin2024efficient} mitigate repeated accesses to identical KV blocks through system KV cache computation, they still incur additional block copying or masking overhead to maintain beam-sequence continuity. 


\begin{figure}[htbp]
    \centering
    \includegraphics[width=0.9\linewidth]{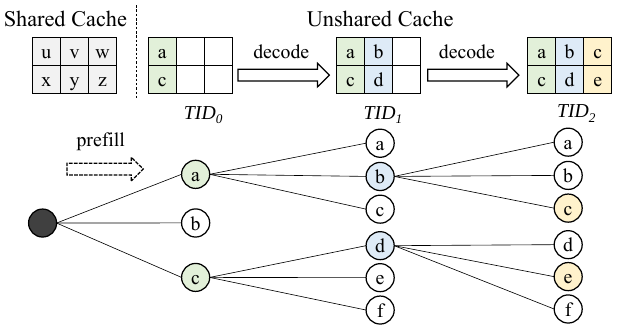}
    \caption{KV cache management in xAttention.}
    \label{fig:cache}
\end{figure}

Figure~\ref{fig:cache} illustrates the KV cache management throughout the whole inference process. The \textit{shared cache} stores the token-level outputs of the prompt produced during the prefill phase, whereas the \textit{unshared cache} holds the Top-$K$ token IDs generated at each decode phase. Specifically, since the number of decode phases ($ND$) is known in advance, xAttention initializes the unshared cache size to exactly the product of $K$ and $ND$, thereby avoiding redundant block loading. In addition, the unshared cache is also managed at the token granularity without extra alignment, avoiding the overhead of block copying. After each decode phase, the beam prefixes are updated and extended with new token IDs.

\begin{figure}[htbp]
    \centering
    \includegraphics[width=1.0\linewidth]{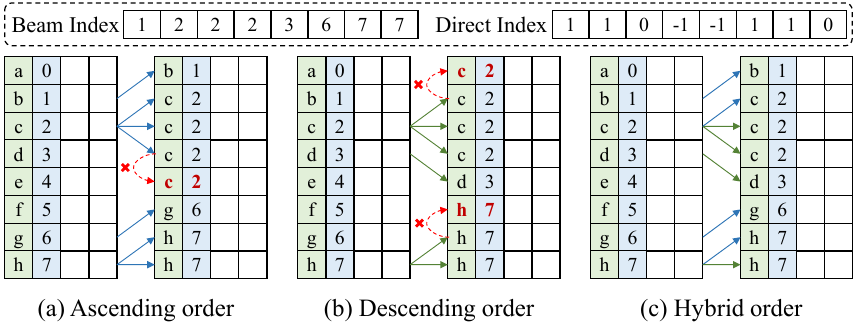}
    \caption{In-place block updates in the unshared cache.}
    \label{fig:select}
\end{figure}

In managing the unshared cache, achieving in-place block updates without introducing additional copies remains challenging. This require only a single memory buffer to modify the blocks, while preserving the continuity of KV loading during attention computation. As shown in Figure~\ref{fig:select}, xAttention updates block contents based on beam indices. However, if the block is naively updated in a strictly downward or upward order, some entries would be overwritten before they are read. To this end, xAttention introduces \textit{direct indices }that indicate the direction: ``1'' for upward writes and ``-1'' for downward writes. Based on these indices, xAttention first performs all upward writes in downward order, and then completes the remaining writes in upward order. This approach eliminates write-before-read hazards and thus ensures correctness.



\subsection{Staged Computation Allocation}
\label{subsec:allocation}

Since the shared and unshared caches do not interfere with each other, xAttention divides the attention computation with common prefixes into a shared stage and an unshared stage. To maximize parallelism, xAttention computes the local attention scores and statistics (i.e., maxima and sums) for the two stages independently. The merging stage applies \textit{OnlineSoftmax} to produce the final logits, followed by \textit{post-processing} to generate the outputs. Inspired by FlashAttention~\cite{dao2022flashattention}, we decompose the computation into tiles and distribute them across hardware units.

\begin{figure}[htbp]
    \centering
    \includegraphics[width=1.0\linewidth]{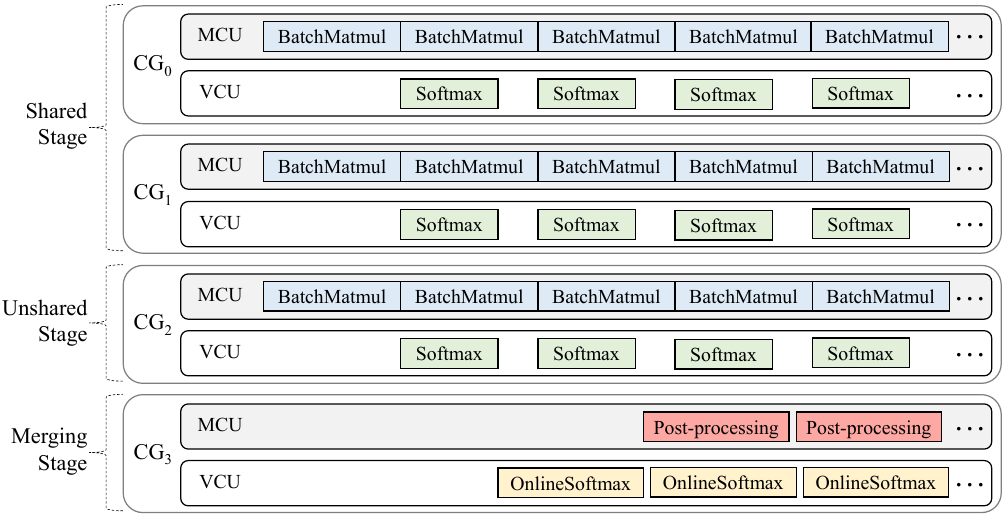}
    \caption{Pipeline parallelism across hardware units.}
    \label{fig:allocation}
\end{figure}

Figure~\ref{fig:allocation} illustrates the pipeline parallelism of attention computation across hardware hierarchies. To improve data locality and avoid frequent cross-CG synchronization, each of the three stages exclusively occupies a CG during execution. In both shared and unshared stages, computations from different tiles are fused using \textit{batchmatmul}, with the \textit{batchmatmul} and \textit{Softmax} operations assigned to the MCU and VCU, respectively. Because the \textit{Softmax} depends on the results of the \textit{batchmatmul}, the MCU and VCU within each CG operate in a pipelined parallel manner. However, different CGs do not have interdependencies and can therefore execute in an embarrassingly parallel fashion. Regarding the merging stage, \textit{OnlineSoftmax} and \textit{Post-processing} are deployed on the MCU and VCU, respectively. Since the merging stage depends on the partial logit results generated by \textit{Softmax}, it forms a pipelined execution across CGs with the preceding stages. The merging stage employs a soft-synchronization mechanism, setting flags in the workspace so that the CG assigned to this stage spin-waits until the required data becomes available.

Given that accelerators typically contain dozens of CGs, effectively partitioning CGs across stages to achieve load balance is critical. On the one hand, the computational volume of the shared stage is generally greater than that of the unshared stage, potentially requiring more CGs. On the other hand, the shared cache only needs to be loaded once, whereas the unshared cache must be updated at each decode step. The limited CGs allocated to the unshared stage exacerbate intra-CG cache thrashing and scratchpad memory conflicts.
As for the merging stage, the optimal CG allocation is strongly correlated with the preceding stages and cannot be predetermined. To this end, xAttention employs a lightweight decision tree regressor~\cite{chen2016xgboost} to predict the performance of each CG partition setting represented as triplets. In addition to the partition setting, the input parameters also include the lengths of unshared and shared caches, i.e., the number of tokens. Note that in mature deployments, GR parameters such as $BW$ and $K$, as well as attention parameters like head size, are fixed. Therefore, there is no need to include them as inputs to the regressor. Consequently, the overall cost of training set collection and model training is feasible.



\section{Beam Search}
\label{sec:beam}

In the decode phase, each beam selects token IDs with Top-$BW$ probabilities from the vocabulary to expand its sequence. Then, all beam sequences are traversed to obtain the tokens with Top-$K$ probabilities to enter the next decode phase. 
We introduce the following optimization techniques.

\subsection{Valid Path Constraint}
\label{sec:constraint}

\begin{figure}[htbp]
    \centering
    \includegraphics[width=1.0\linewidth]{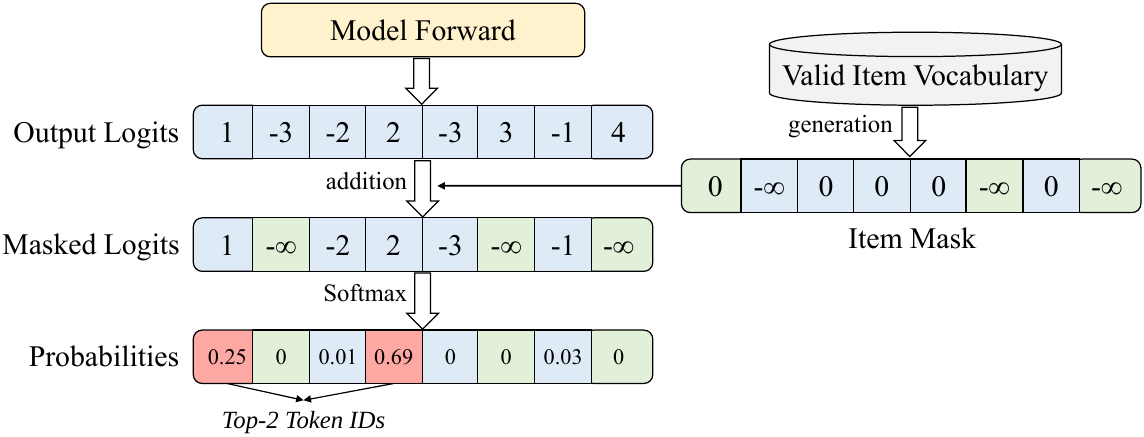}
    \caption{Valid path constraint during token generation.}
    \label{fig:constrained}
\end{figure}

Given the vast combinatorial space of token IDs, not all token-ID triplets correspond to actual items. To address this issue, we borrow the idea of constraint decoding~\cite{dong2024xgrammar} in LLMs and leverage item masks to filter out invalid token IDs. As illustrated in Figure~\ref{fig:constrained}, xBeam generates an item mask based on the pre-built valid item vocabulary and incorporates the mask into the model's output logits through element-wise addition. Once the masked logits are processed by the \textit{Softmax} function, the probabilities of invalid token IDs become vanishingly small, effectively preventing them from being selected.

On the other hand, xBeam employs a combination of sparse and dense storage to improve mask generation efficiency. For example, during the first decode step, each beam typically contains thousands of candidate tokens. To mitigate latency, the mask is stored in a dense format and pre-generated during model loading, eliminating the need for on-demand generation at runtime. In contrast, during the final decode step, each beam only contains few candidate tokens. In this case, xBeam stores the relevant positions in a sparse format and performs in-place updates to the existing mask. Owing to the small extent of these modifications, the incurred runtime generation latency remains within an acceptable range.

\subsection{Early Sorting Termination}
\label{sec:termination}

During token generation, beam search accumulates log-probabilities (\textit{log\_prob}) rather than multiplying raw probabilities to ensure numerical stability. After that, beam search selects the Top-$K$ token sequences from the filtered candidate sequences, which can be viewd as a partial sorting problem. Another feature is that the \textit{log\_prob} results for each beam are inherently in descending order. Based on the above considerations, xBeam adopts an early termination mechanism to improve sorting efficiency.


\begin{figure}[htbp]
    \centering
    \includegraphics[width=1.0\linewidth]{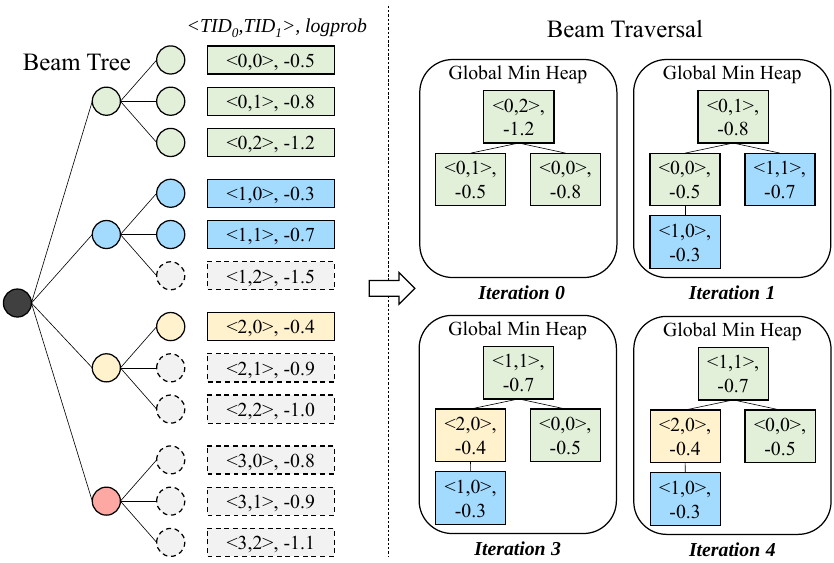}
    \caption{The sorting process with early termination.}
    \label{fig:termination}
\end{figure}

Figure~\ref{fig:termination} shows the sorting process with early termination. xBeam maintains a global min heap of size $K$ to store the top-ranking token sequences along with their associated \textit{log\_prob} values. Specifically, xBeam visits the leaves of each sub-beam tree sequentially. During the traversal of each beam, if the leaf's \textit{log\_prob} exceeds that of the heap's top element, it is inserted into the heap, and the heap structure is adjusted to maintain order. Otherwise, the sorting operation of that beam is terminated immediately. Finally, xBeam retrieves the top-$K$ token sequences. In this way, xBeam reduces sorting overhead while ensuring high-quality sequence selection.




\subsection{Data Structure Reuse}
\label{sec:data}

Beam search typically requires a larger $K$ (e.g., 512) to ensure accurate recommendations. However, continuously generating new candidate sequences while discarding old ones leads to substantial waste of computational and memory resources. This is because the frequent creation and destruction of data structures, as well as the associated data movement operations, incur significant overhead. Fortunately, since the $K$ is predetermined and fixed throughout the beam search process, this provides an opportunity for resource reuse. Specifically, during the generation of new candidate sequences, xBeam does not allocate entirely new data structure; instead, it reuses the data structure previously occupied by old sequences. Once beam search completes, xBeam updates the memory regions corresponding to the old request with the content of the new incoming request. 


\section{System Scheduling}
\label{sec:system}

\begin{figure}[htbp]
    \centering
    \includegraphics[width=1.0\linewidth]{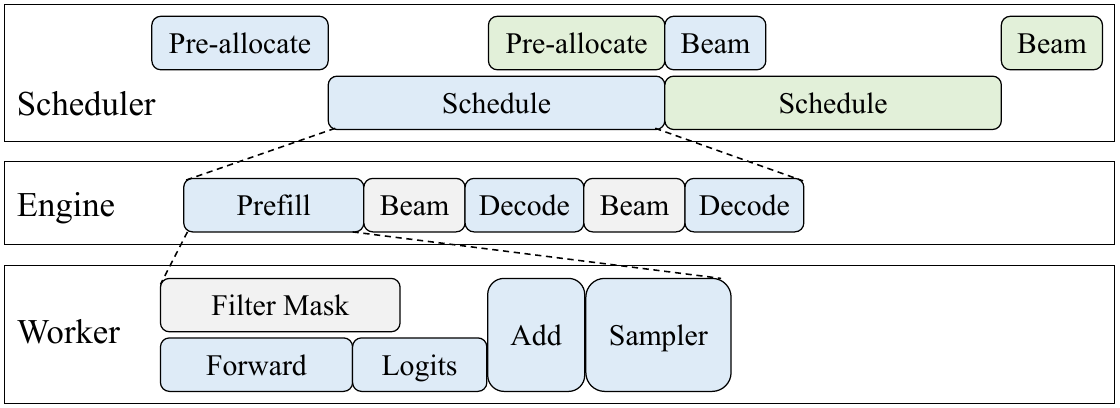}
    \caption{The overall pipeline of xSchedule.}
    \label{fig:pipeline}
\end{figure}

Figure~\ref{fig:pipeline} shows the overall pipeline of xSchedule. First, the scheduler pre-allocates resources and prepares the necessary embedding information on the host side. Once prepared, the scheduler initiates the corresponding phases and hands them over to the engine for processing. Then, the engine manages the continuous execution of one prefill followed by several beam search and decode combinations. Note that beam search and decode are tightly coupled, leaving no room for pipelining across phases. The worker is responsible for executing all tasks associated with a specific phase (e.g., prefill).

Such a multi-level pipeline offers numerous opportunities for system-level optimization to improve overall execution efficiency. For example, the host-side scheduler concurrently processes requests by overlapping control and computation flows. Specifically, the \textit{Schedule} involves device-side control (e.g., kernel dispatch), which can overlap with the \textit{Beam} and \textit{Pre-allocate} computations. Moreover, xSchedule captures a series of device-side operations (e.g., kernels and memory copies) in the form of a graph and submits them all at once. This approach significantly reduces kernel launch overhead and enables execution without intervention. On the other hand, xSchedule strives to overlap host and device operations as much as possible. The filter mask generation is performed on the host side, overlapping with the device-side model forward pass that produces logits. In addition, the H2D transfer of the mask is performed concurrently with attention computations.


In addition to pipeline optimizations, fully exploiting the accelerator capabilities is also crucial for improving request throughput. xSchedule employs a dynamic batching strategy similar to that used in LLMs, aggregating incoming requests into larger batches. However, in recommendation scenarios, the request concurrency scale can vary drastically, and individual requests may contain tens to thousands of tokens. To handle such variability, xSchedule automatically adjusts the batch size based on the token capacity. Meanwhile, the batching interval is constrained by the SLO: if the waiting delay reaches the allocated quota, the batch is dispatched for computation immediately. Furthermore, given that GR models contain far fewer parameters than LLMs and therefore have lower computational load, sequentially executing batches may still leave certain units underutilized in the spatial dimension. To this end, xSchedule employs a multi-stream strategy to process batches concurrently, where each stream independently handles requests within a single batch. 

\section{Implementation}
\label{sec:implementation}

We implement xGR on top of xLLM~\cite{liu2025xllm}, comprising 17k lines of Ascend C/CUDA/C++ code and 4k lines of Python code. Notably, the system and operator designs are portable across Ascend NPU and NVIDIA GPU, as modern accelerators generally feature multi-level memory hierarchies and heterogeneous compute units. These can be uniformly abstracted to efficiently support matrix- and vector-related operations. However, the specialized components lead to differences in fine-grained optimizations. For example, On NVIDIA GPU, xGR leverages the tensor memory accelerator (TMA) of the Hopper architecture to load tiles from global memory into shared memory. This is made possible by our novel KV cache management that ensures contiguous memory layout. In this case, TMA can perform asynchronous DMA transfers in a single operation, maximizing throughput while reducing the overhead of address calculations and load/store instructions.



%


\section{Evaluation}
\label{sec:evaluation}

\subsection{Experimental Setup}
\label{sec:setup}











\textbf{Hardware.} We evaluate our system on an Ascend cluster and a GPU cluster. The Ascend cluster comprises 64 nodes, each with 16 Ascend NPUs (64GB) interconnected via HCCS. The GPU cluster comprises 8 nodes, each with 8 NVIDIA H800 GPUs (80GB) interconnected via NVLink.


\textbf{Models.} We select Qwen3 \cite{yang2025qwen3} and OneRec \cite{kong2025minionerec}, widely adopted as representative models for GR tasks. 

\textbf{Baselines.} We compare our system with two baselines including vLLM \cite{kwon2023efficient} and xLLM \cite{liu2025xllm}. vLLM is a high-throughput serving engine using PagedAttention. We use the vllm-ascend porting for the Ascend cluster. xLLM is an industrial inference framework optimized for throughput. It leverages PagedAttention's mechanism for memory management, providing efficient support for both platforms. During the beam search phase, we set $BW$ and $K$ to be equal, which aligns with production scenarios without sacrificing generality.



\textbf{Datasets.} We use two datasets including Amazon Review \cite{hou2024bridging} and JD Trace. Amazon Review is a public dataset for benchmarking recommendation tasks. JD trace is taken from production scenarios, featuring dynamic traffic patterns. 

\textbf{Metrics.} We focus on two critical metrics for user experience: average latency and P99 latency under varying RPS.


\subsection{End-to-end Performance}
\label{sec:end}

\begin{figure}[!t]
    \centering
    \begin{minipage}[b]{1.02\linewidth}
        \centering
        \footnotesize
        \includegraphics[width=\linewidth]{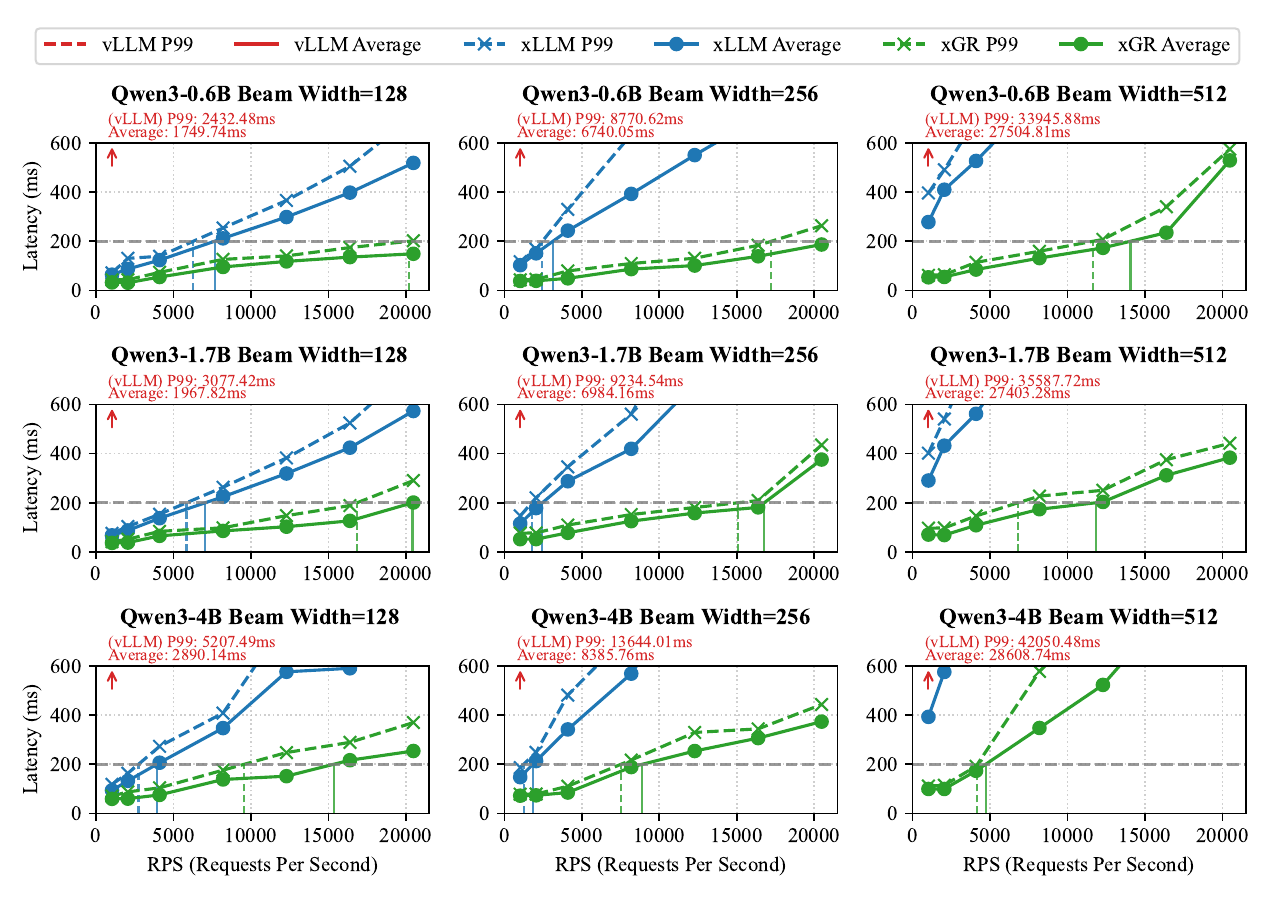}
        (a) Amazon Review Dataset
        \label{fig:qwen3_amazon}
    \end{minipage}
    
    \vspace{0.2cm} 
    
    \begin{minipage}[b]{1.02\linewidth}
        \centering
        \footnotesize
        \includegraphics[width=\linewidth]{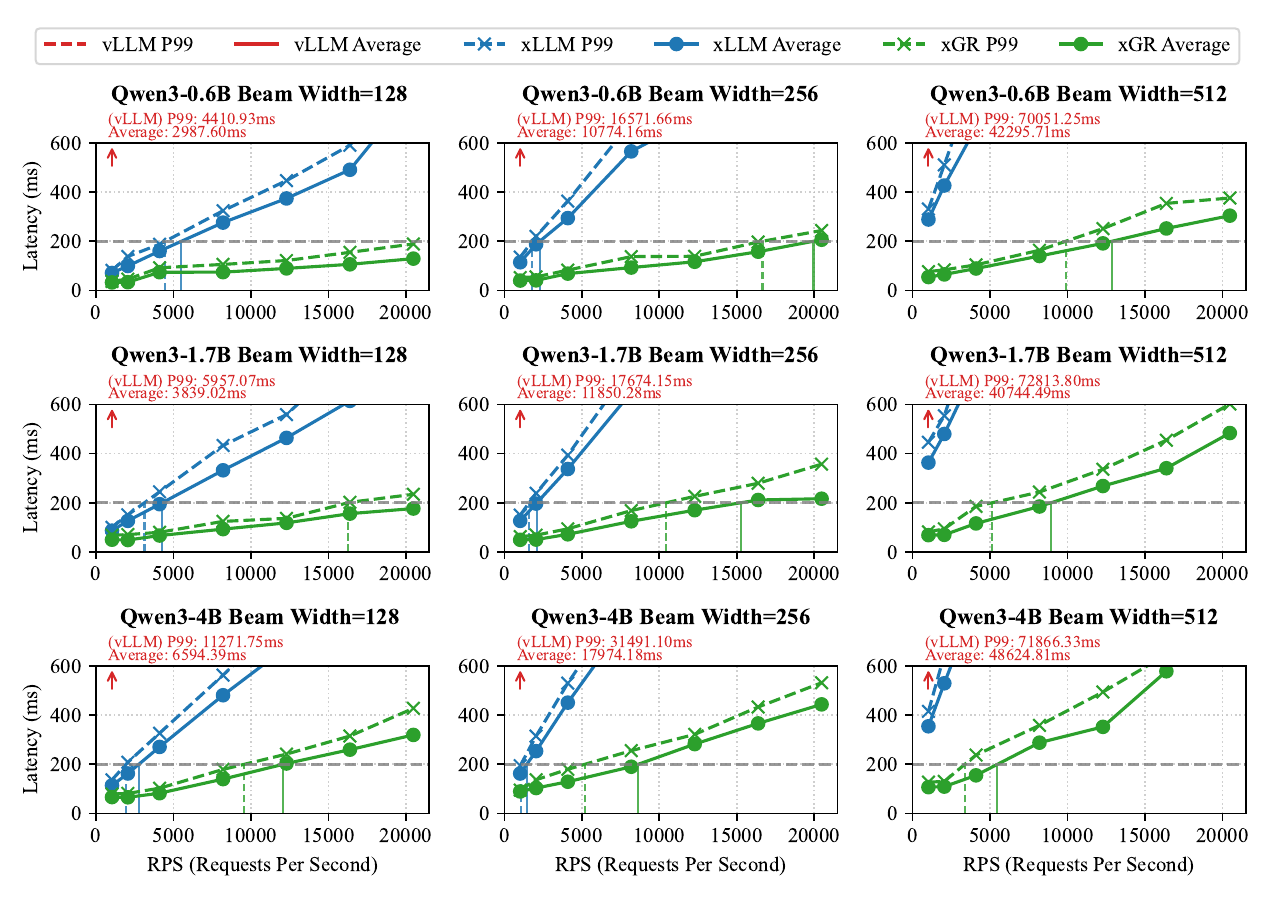}
        (b) JD Dataset
        \label{fig:qwen3_JD}
    \end{minipage}
    
    \caption{End-to-end comparison of Qwen3 models on the Ascend cluster using (a) Amazon Review dataset and (b) JD dataset.}
    
    \label{fig:qwen3_combined}
\end{figure}
\begin{figure}[!t]
    \centering
    \begin{minipage}[b]{1.02\linewidth}
        \centering
        \footnotesize
        \includegraphics[width=\linewidth]{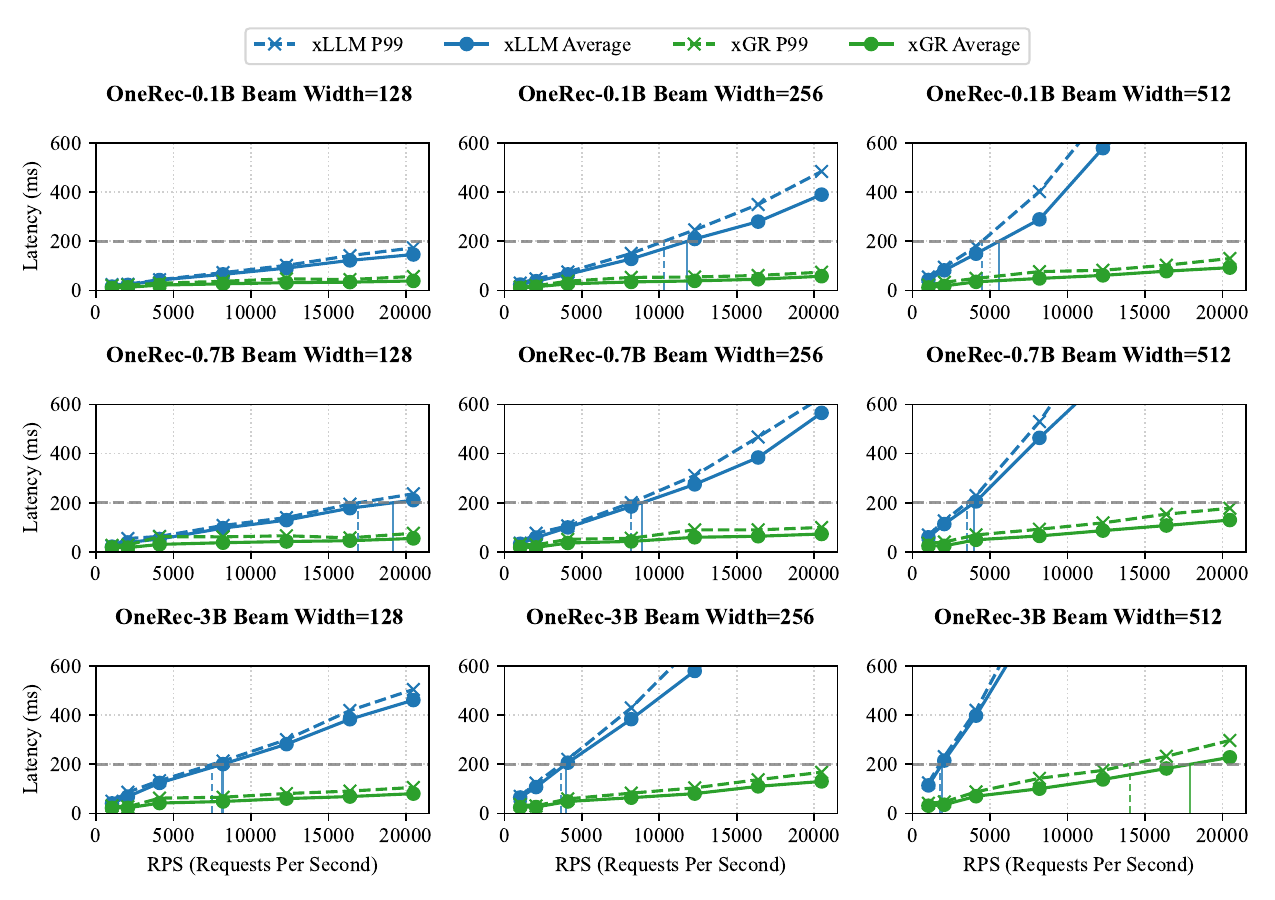}
        (a) Amazon Review Dataset
        \label{fig:onerec_amazon}
    \end{minipage}
    
    \vspace{0.2cm} 
    
    \begin{minipage}[b]{1.02\linewidth}
        \centering
        \footnotesize
        \includegraphics[width=\linewidth]{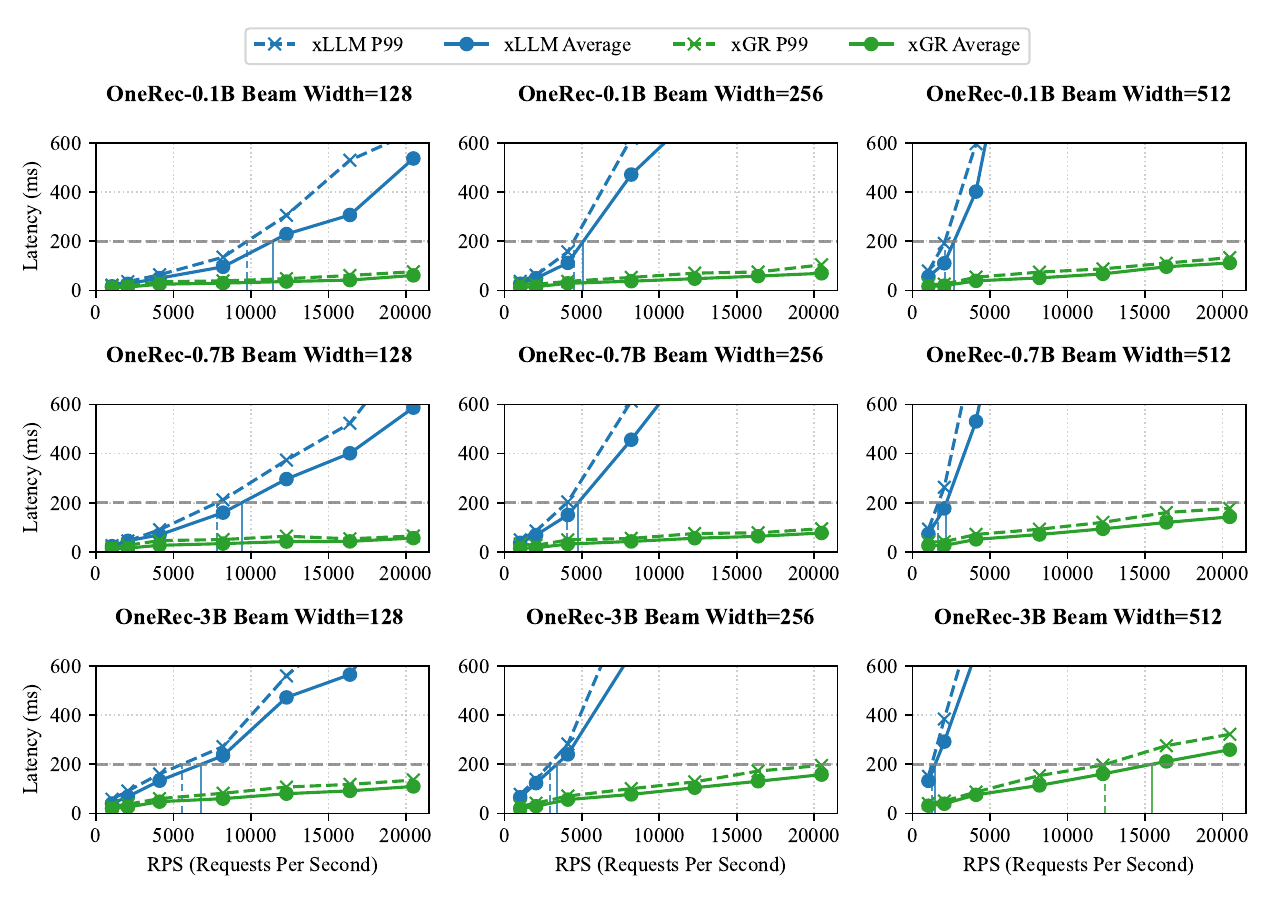}
        (b) JD Dataset
        \label{fig:onerec_JD}
    \end{minipage}
    
    \caption{End-to-end comparison of OneRec models on the Ascend cluster using (a) Amazon Review dataset and (b) JD dataset.}
    \label{fig:onerec_combined}
\end{figure}

We conduct the experiments on the Ascend cluster using Qwen3 model (ranging from 0.6B to 4B parameters) and the OneRec model (ranging 0.1B to 3B parameters).
Figure \ref{fig:qwen3_combined} and Figure \ref{fig:onerec_combined} illustrate the curves for Qwen3 and OneRec, respectively. Note that vLLM does not natively support OneRec and is thus not included in Figure \ref{fig:onerec_combined}. xGR consistently outperforms both baselines across all cases. 
Specifically, under strict latency constraints (P99 $\le 200$ms), xGR sustains a significantly higher throughput, exceeding xLLM by at least 2.89$\times$ and 8.05$\times$ on Qwen and OneRec models, respectively.
As the RPS increases, the baselines experience a sharp rise in latency and quickly reaches the constraint. In contrast, xGR maintains a relatively smooth latency increase even at high loads, demonstrating the efficacy of staged attention computation and KV cache separation.



GR heavily relies on large beam widths to ensure retrieval diversity. The experiments under varying beam widths show that the performance gap widens significantly as the beam width increases. For small beam width (128), the baselines struggle to meet SLO requirements even at moderate request concurrency. At larger beam widths (e.g., 512), the baselines fail to scale effectively, with latency increasing non-linearly. Conversely, xGR leverages the high NPU parallelism for both sorting and filtering, coupled with shared prefix management. This design allows xGR to exhibit only a slight increase in latency even when the beam width grows by multiples.



We further verify the scalability of xGR across different model scales. xGR consistently delivers superior performance across the spectrum. Notably, on larger models (e.g., Qwen3-4B), the memory bandwidth savings become even more pronounced. By avoiding the redundant loading of shared prompt contexts for varying beams, xGR supports high concurrency robustly, whereas the baselines frequently trigger preemption or recomputation under similar conditions. Since vLLM-Ascend is not well optimized, its redundant memory accesses under large-width beam search lead to severe underutilization of compute units. We compare xGR only with xLLM for Ascend NPU in the following experiments.



%

\begin{figure}[t]
    \centering
    \footnotesize

    \noindent
    \begin{minipage}[t]{0.4\textwidth}
        \centering
        \includegraphics[width=\linewidth]{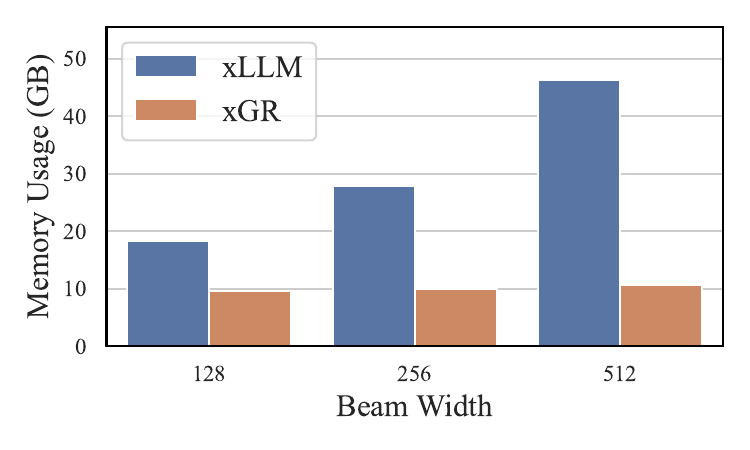}
        (a) Fixed Input Length
    \end{minipage}\hfill%
    \begin{minipage}[t]{0.4\textwidth}
        \centering
        \includegraphics[width=\linewidth]{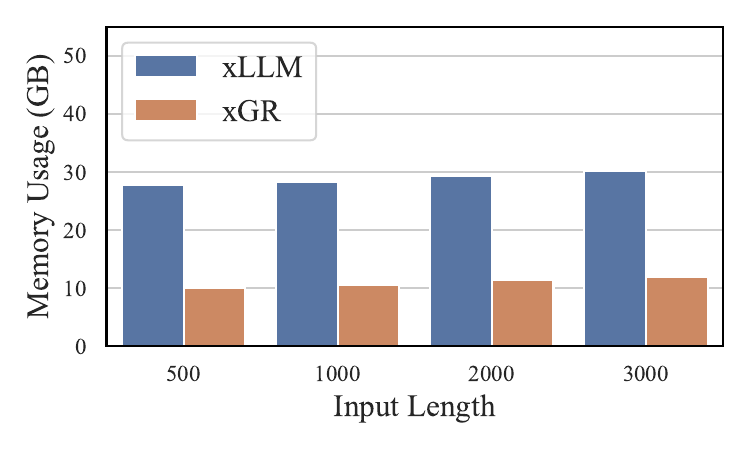}
        (b) Fixed Beam Width
    \end{minipage}

    \caption{Peak memory usage comparison between xLLM and xGR.}
    \label{fig:memory_efficiency}
\end{figure}

\subsection{Memory Efficiency}
\label{sec:memory_Efficiency}


We analyze the memory usage of xGR compared to xLLM using the Qwen3-4B model on a single Ascend NPU, where the RPS is set to 4.
Figure \ref{fig:memory_efficiency}(a) illustrates the results as the beam width increases from 128 to 512 with a fixed input lengh of 1k tokens. While xGR maintains a stable memory footprint (around 10GB), xLLM exhibits super-linear growth. Specifically, xLLM consumes a 46.3GB compared to xGR's 10.6GB with $BW=512$. The excessive memory consumption in xLLM stems from the additional block copying inherent in PagedAttention, leading to severe redundancy and fragmentation. xGR addresses this through separated KV cache management, maintaining a single physical copy of the shared prefix and just enough space to store the decoded tokens.





We further evaluate memory scalability under varying input lengths, as shown in Figure~\ref{fig:memory_efficiency}(b). 
With the $BW$ fixed at 256, xGR maintains a remarkably low footprint—peaking at only 12.0GB even with 3k tokens—significantly outperforming xLLM, which consumes around 30GB. This confirms that xGR decouples memory usage from sequence length, ensuring robustness for long-context recommendation tasks.

    


\begin{figure*}[t]
    \centering
    \footnotesize

    \noindent
    \begin{minipage}[t]{0.32\textwidth}
        \centering
        \includegraphics[width=\linewidth]{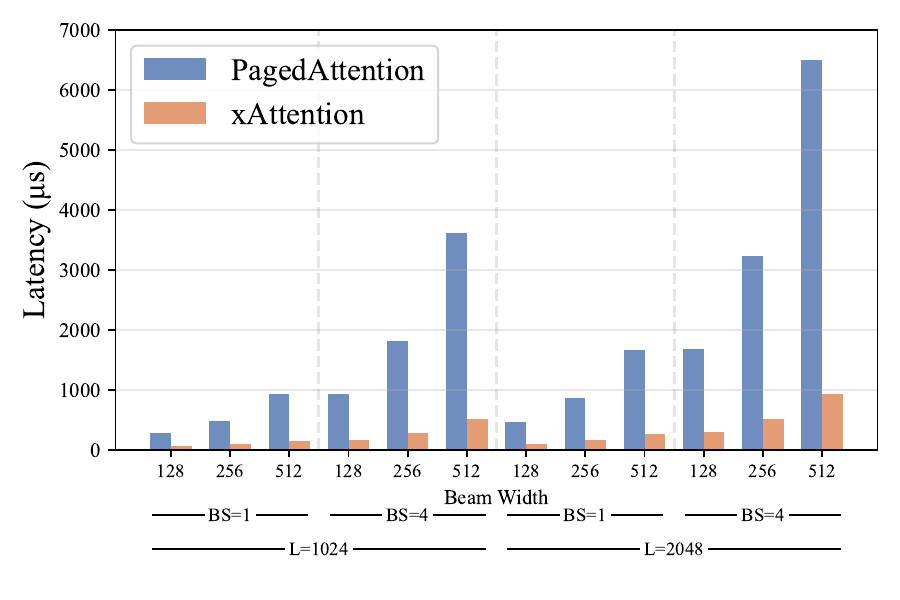}
     
        (a) Kernel Latency
    \end{minipage}\hfill%
    \begin{minipage}[t]{0.32\textwidth}
        \centering
        \includegraphics[width=\linewidth]{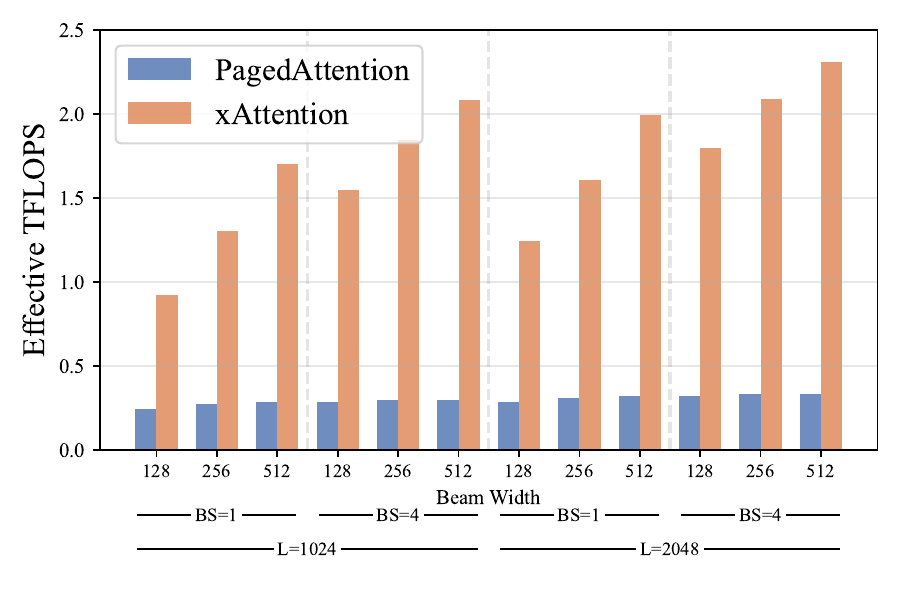}
        
        (b) Computational Throughput
    \end{minipage}\hfill%
    \begin{minipage}[t]{0.32\textwidth}
        \centering
        \includegraphics[width=\linewidth]{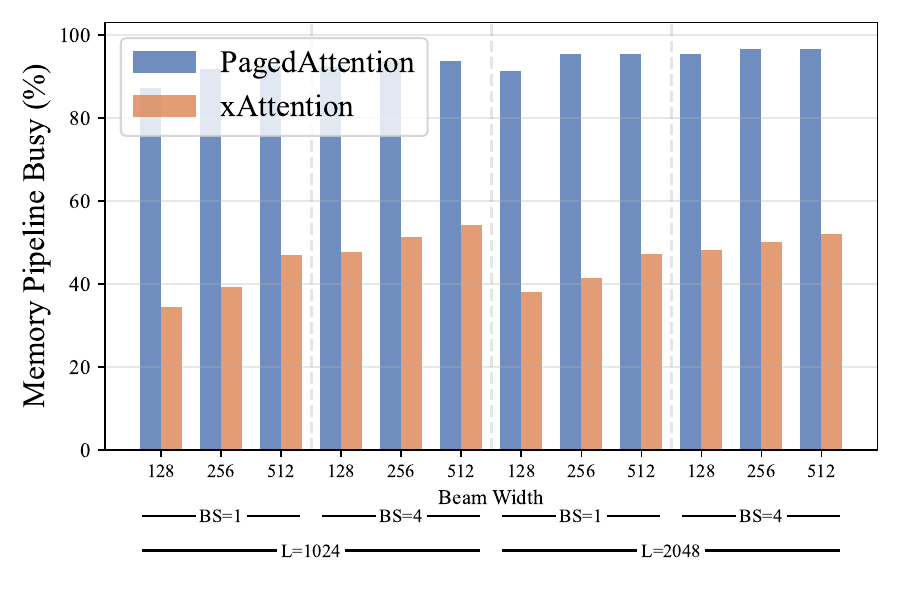}
        
        (c) Memory Access Overhead
    \end{minipage}

    \caption{Kernel efficiency of PagedAttention and xAttention, where $BS$ and $L$ refer to batch size and input length, respectively.}
    \label{fig:kernel_utilization}
\end{figure*}

%
%
%
%
%

\subsection{Kernel Efficiency}
\label{sec:kernel}



We analyze fine-grained kernel efficiency on Ascend NPU across varying input lengths, beam widths and batch sizes. As shown in Figure \ref{fig:kernel_utilization}, we focus on three key metrics including kernel latency, computational throughput, and memory access overhead. Regarding latency and throughput, xAttention demonstrates superior performance, particularly at large beam width (e.g., 512). It reduces kernel latency by approximately 6.6$\times$ and boosts computational throughput by 7$\times$. This indicates that xAttention effectively saturates the compute units, whereas PagedAttention suffers from idle stalls.


We also perform memory profiling via Ascend NPU's Profiler. As show in Figure~\ref{fig:kernel_utilization}(c), PagedAttention is heavily memory-bound. The memory access pipeline spends an average of 93.4\% of its time in a busy state. This is primarily due to repeated memory load to the same KV cache prefix. In contrast, xAttention maintains a steady busy rate of around 52\%, successfully transforming the workload from memory-bound to compute-bound and maximizing hardware utilization.

\begin{figure*}[ht]
    \centering
    \includegraphics[width=1.0\linewidth]{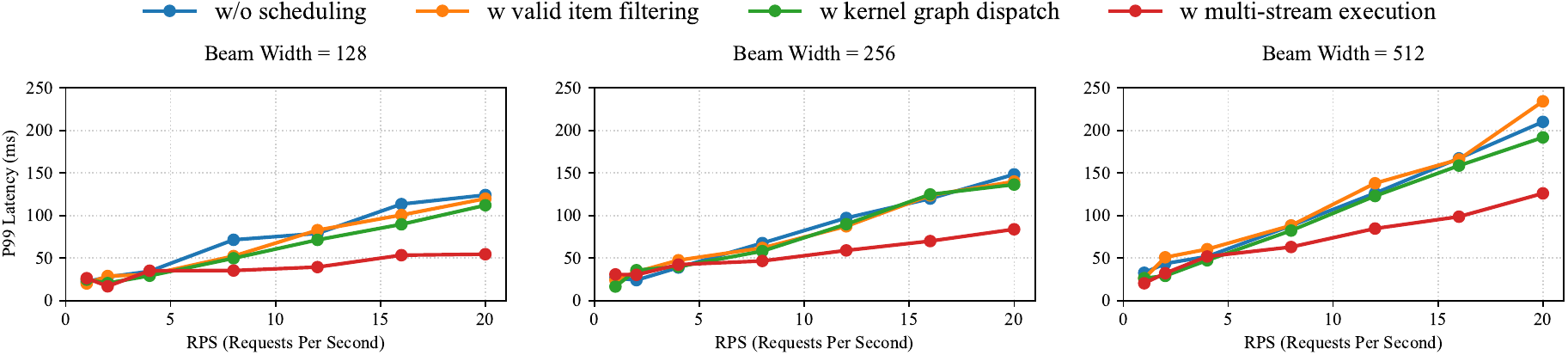}
    \caption{Ablation study of scheduling optimizations using OneRec-0.1B on the Amazon Review dataset.}
    \label{fig:ablation}
\end{figure*}

\subsection{Scheduling Analysis}
\label{sec:scheduling}





To quantify the individual contributions of xSchedule, we conduct an ablation study using the OneRec 0.1B model on the Amazon Review dataset. We establish the xGR baseline that implements the core xAttention and xBeam components but lacks scheduling optimizations. We enable valid item filtering, kernel graph dispatch, and multi-stream execution separately to evaluate their individual impacts. As observed in Figure \ref{fig:ablation}, without multi-stream optimization, xGR suffers from serial execution overhead and fails to overlap H2D data transfer with device-side computation. The kernel graph dispatch optimization allows xGR to capture the sequence of kernels into a static graph structure, drastically reducing the CPU interactions required per decode phase.

We also evaluate the overhead of ensuring recommendation validity via item filtering. Figure \ref{fig:ablation} demonstrate that the filtering overhead is negligible. This efficiency stems from the fully device-resident implementation, ensuring that validity checks are performed within the beam search kernel. 


\subsection{Deployment on NVIDIA GPUs}
\label{sec:nvidia}
\begin{figure}[!ht]
    \centering
    \includegraphics[width=0.9\linewidth]{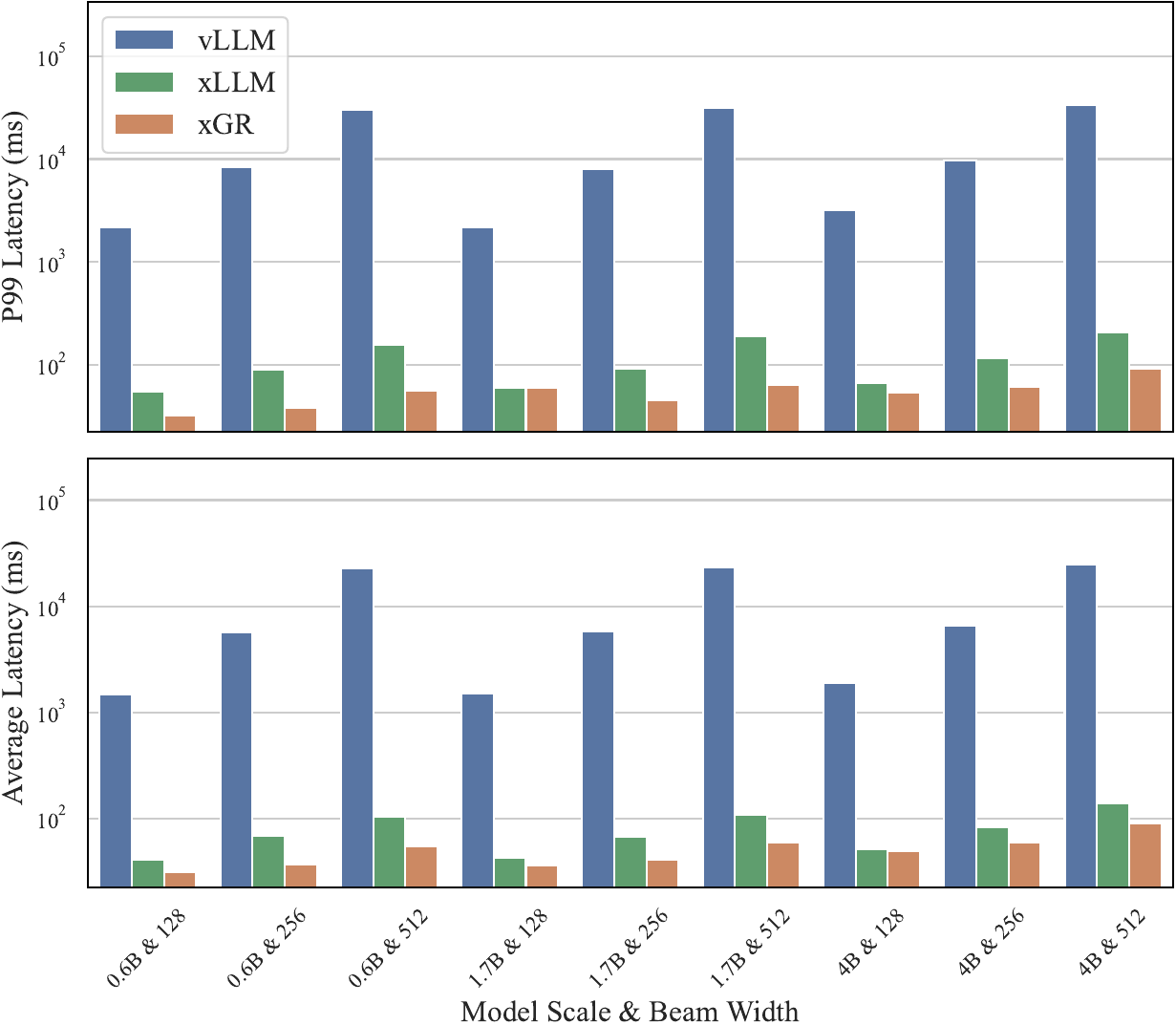}
    \caption{End-to-end comparison on the GPU cluster using the Amazon Review dataset at a fixed RPS of 64.}
    \label{fig:gpu}
\end{figure}

To demonstrate the portability of xGR, we extend the evaluation to the GPU cluster. 
While the architectural details of NVIDIA GPU differ from that of Ascend NPU, the fundamental challenges such as ``memory wall'' and KV cache management induced by large beam widths remain consistent.
Figure \ref{fig:gpu} presents the P99 and average latency comparison across various model scales (0.6B, 1.7B, 4B) and beam widths ($B \in \{128, 256, 512\}$). The results on the GPU cluster mirror the trends observed on the Ascend NPU. xGR consistently achieves significantly lower latency and higher sustainable throughput. Notably, despite H800 GPU's high memory bandwidth and H2D bandwidth (employs PCIe Gen5), as well as large computational capacity, it still performs poorly on vLLM and xLLM. This indicates that hardware enhancement alone is insufficient to effectively address the unique challenges of GR, requiring a comprehensive system redesign.

\section{Related Work}
\label{sec:related}

\subsection{DLRM Serving System}


DLRM serving must meet strict latency requirements on the order of hundreds of milliseconds, placing stringent demands on inference efficiency~\cite{sima2022ekko,ke2022hercules,wei2022gpu,ye2023grace,yang2025gpu}.
Ekko~\cite{sima2022ekko} exploits the temporal locality to prioritize the sending of memory updates.
GRACE~\cite{ye2023grace} introduces item co-occurrence graph that records co-accessed item combinations.
Prism~\cite{yang2025gpu} schedules CPU/GPU-intensive subgraphs to disaggregated nodes.
However, the aforementioned studies remain constrained by the model scaling law and cascade architecture. 

\subsection{LLM Inference Engine}

The growing demand for LLM serving has driven the development of inference frameworks~\cite{yu2022orca,kwon2023efficient,zhong2024distserve,zheng2024sglang,patel2024splitwise,zhu2025nanoflow}. 
vLLM~\cite{kwon2023efficient} offers LLM execution beyond GPU memory capacity, which leads to larger batch sizes. 
xLLM~\cite{liu2025xllm} adopts a workload-adaptive dynamic prefill-decode (PD) disaggregation policy for instance scheduling.
PrefillOnly~\cite{du2025prefillonly} assumes that the model outputs a single token, which enables predictable job completion time (JCT) and KV-cache storage. However, such aggressive optimization is misaligned with production scenarios where each decode phase also incurs substantial computation.
Tokasaurus~\cite{juravsky2025tokasaurus,juravsky2024hydragen} targets high-throughput decoding by performing shared prefix computation on a batch of sequences.
Although its core idea is similar to xAttention, the coarse-grained paged management still results in additional block copying, especially when dealing with amounts of invalid tokens during beam search. We will dedicate efforts to integrating it into GR for further evaluation in the future.



There are also lines of work that focus on memory management (e.g., KV cache)~\cite{guo2024gmlake,yao2025cacheblend,yu2025ic,zhang2025diffkv}. 
IC-Cache~\cite{yu2025ic} selectively evicts examples to preserve those that yield the most effective responses.
DiffKV~\cite{zhang2025diffkv} packs fragmented free memory lists into contiguous regions on GPU. 
A recent work, BAT\cite{sun2026bat} selects either the user or item as the prompt prefix for efficient KV cache reuse.
However, it only applies to the cascade architecture that replaces the ranking stage with an LLM. End-to-end GR inference directly uses the user interaction history as input, making prefix cache separation unavailable.


\section{Conclusion}
\label{sec:conclusion}

In this paper, we propose xGR, an efficient GR serving system that meets low-latency requirements under high request concurrency. First, we implement an attention operator that separates KV cache management and employs staged computation allocation. Second, we introduce a beam search algorithm that filters invalid items and enables early search termination. Third, we design a scheduling mechanism that allows pipeline parallelism throughout the entire system. The experiments with real-world datasets demonstrate that xGR achieves at least 2.89$\times$ throughput compared to the state-of-the-art baseline under strict latency constraints.
%

%

\bibliographystyle{plain}
\bibliography{main}

\end{document}